\documentclass{article}
\usepackage[preprint]{neurips_2026}
\usepackage[T1]{fontenc}
\usepackage{hyperref}
\usepackage{url}
\usepackage{booktabs}
\usepackage{amsfonts}
\usepackage{amsmath}
\usepackage{amssymb}
\usepackage{nicefrac}
\usepackage{microtype}
\usepackage{xcolor}
\usepackage{graphicx}
\usepackage{subcaption}
\usepackage{multirow}
\usepackage{makecell}
\usepackage{tcolorbox}

\newcommand{\crossmem}{cross-scene affordance memory}
\newcommand{\inscmem}{in-scene spatial memory}
\newcommand{\Crossmem}{Cross-Scene Affordance Memory}
\newcommand{\Inscmem}{In-Scene Spatial Memory}
\newcommand{\gemmask}{geometry memory mask}

\title{Grounding by Remembering: Cross-Scene and In-Scene Memory for 3D Functional Affordances}

\author{
Qirui Wang$^{1}$,
Jingyi He$^{1}$,
Yining Pan$^{2}$,
Xulei Yang$^{2}$,
Shijie Li$^{2}$\\
$^{1}$TUM \\
$^{2}$A*STAR
}

\begin{document}
\maketitle

% ─────────────────────────────────────────────────────────────────────────────
\begin{abstract}
% ─────────────────────────────────────────────────────────────────────────────
Functional affordance grounding requires more than recognizing an
object: an agent must localize the specific region that supports an
interaction, such as the handle to pull or the button to press.
This is difficult for training-free vision-language pipelines because
actionable regions are often small, visually ambiguous, and repeated
across multiple same-category instances in a scene.
We propose \textsc{AffordMem}, a framework that grounds 3D functional
affordances by remembering geometry at two levels.
The first is cross-scene affordance memory: the agent
maintains a category-level memory bank of RGB images with affordance
regions rendered as overlays, and recalls the most informative examples
at query time to guide a frozen VLM toward small operable subregions
that text-only prompting consistently misses.
The second is in-scene spatial memory: as the agent
processes the scene, it organizes candidate instances and their 3D
spatial relations into a structured scene graph, enabling the language
model to resolve references over distant or currently unobserved
candidates such as ``the second handle from the top.''
\textsc{AffordMem} requires no model fine-tuning and no target-scene
annotation, using a reusable memory bank built from source scenes.
On SceneFun3D~\citep{scenefun3d}, our method improves over the prior
training-free state of the art by $+3.23$ AP$_{50}$ on Split~0 and
$+3.7$ AP$_{50}$ on Split~1.
Ablation studies support complementary benefits: \crossmem{} improves
fine-grained localization, while \inscmem{} provides the larger gain on
spatially qualified queries. The project homepage is available at this \href{https://sj-li.com/PROJ/AffordMem/}{link}.

\end{abstract}

% ─────────────────────────────────────────────────────────────────────────────
\section{Introduction}
\label{sec:intro}
% ─────────────────────────────────────────────────────────────────────────────

% \begin{figure*}[t]
%   \centering
%     \caption{\textbf{Two core challenges in 3D functional affordance
%     grounding.}
%     \emph{Left}: granularity ambiguity---a VLM returns the entire drawer
%     panel rather than the small graspable handle.
%     \emph{Right}: instance disambiguation---resolving ``the second drawer
%     from the top'' requires global scene context that no single view
%     provides.
%     \textsc{AffordMem} addresses both via \crossmem{} (category-level
%     visual memory for fine-grained localization) and \inscmem{} (structured
%     scene graph for spatially-grounded selection).}
%   \includegraphics[width=\linewidth]{figures/teaser.png}
%   \label{fig:teaser}
% \end{figure*}

\begin{figure*}[t]
    \centering
      \caption{\textbf{Two memory failures in training-free 3D affordance grounding.}
  \emph{Left}: local visual grounding suffers from granularity ambiguity,
  often selecting a coarse object region instead of the fine-grained
  operable part. \Crossmem{} mitigates this by recalling cross-scene
  affordance exemplars as visual memory.
  \emph{Right}: spatially qualified queries require resolving both
  instance ambiguity and long-range dependency. \Inscmem{} builds a global
  in-scene spatial memory over candidate parts and reference objects,
  allowing the correct affordance instance to be selected beyond the
  current local view.}\label{fig:motivation}
    \label{fig:overall}    
    \begin{subfigure}[b]{0.48\columnwidth}
        \centering
        \caption{Cross-scene memory for fine-grained grounding}
        \includegraphics[width=\linewidth,trim=155 30 155 30,clip]{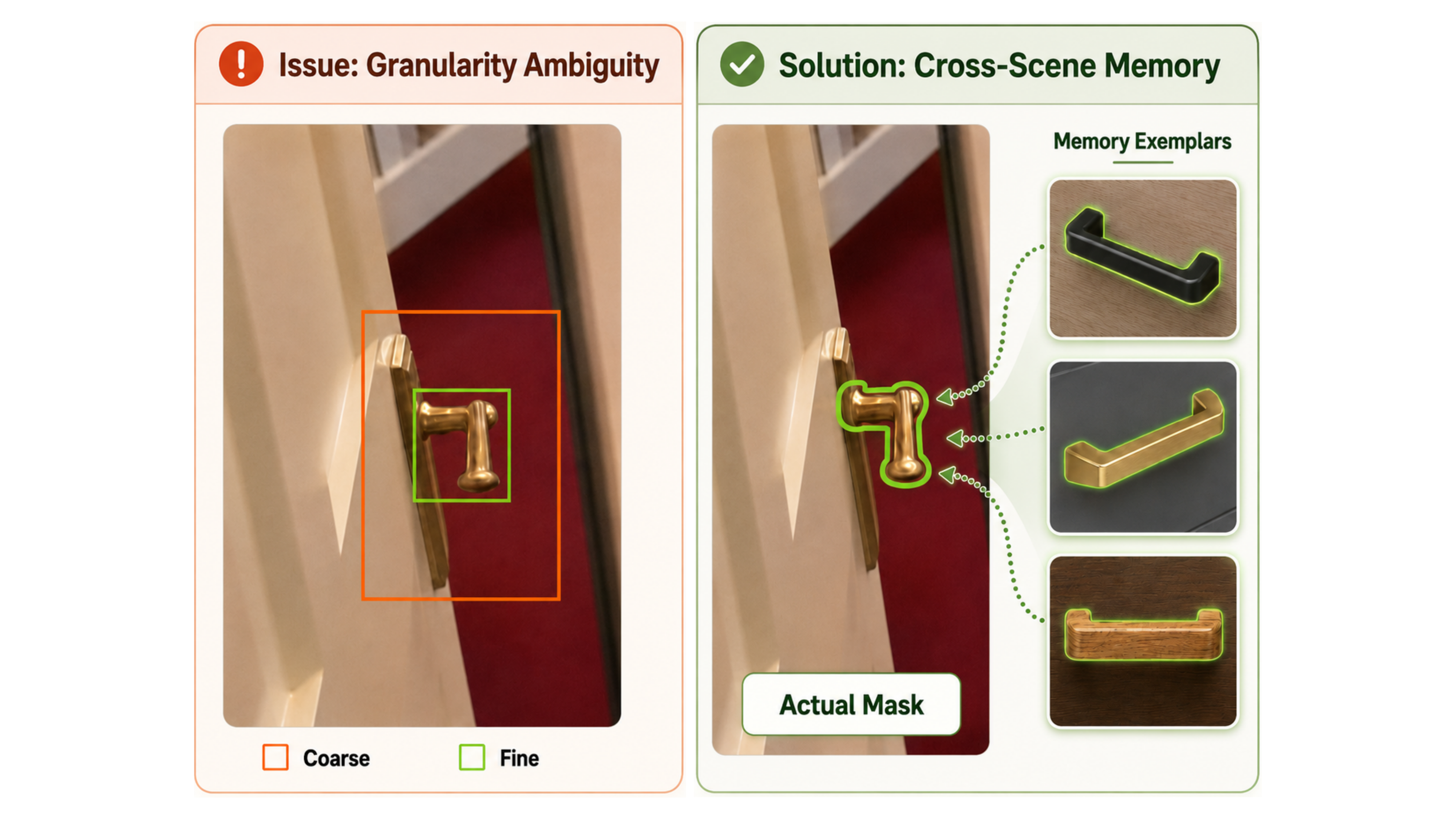}
        \label{fig:left}
    \end{subfigure}
    \hfill
    \begin{subfigure}[b]{0.51\columnwidth}
        \centering
        \caption{In-scene memory for spatial instance selection}
        \includegraphics[width=\linewidth,trim=90 25 90 25,clip]{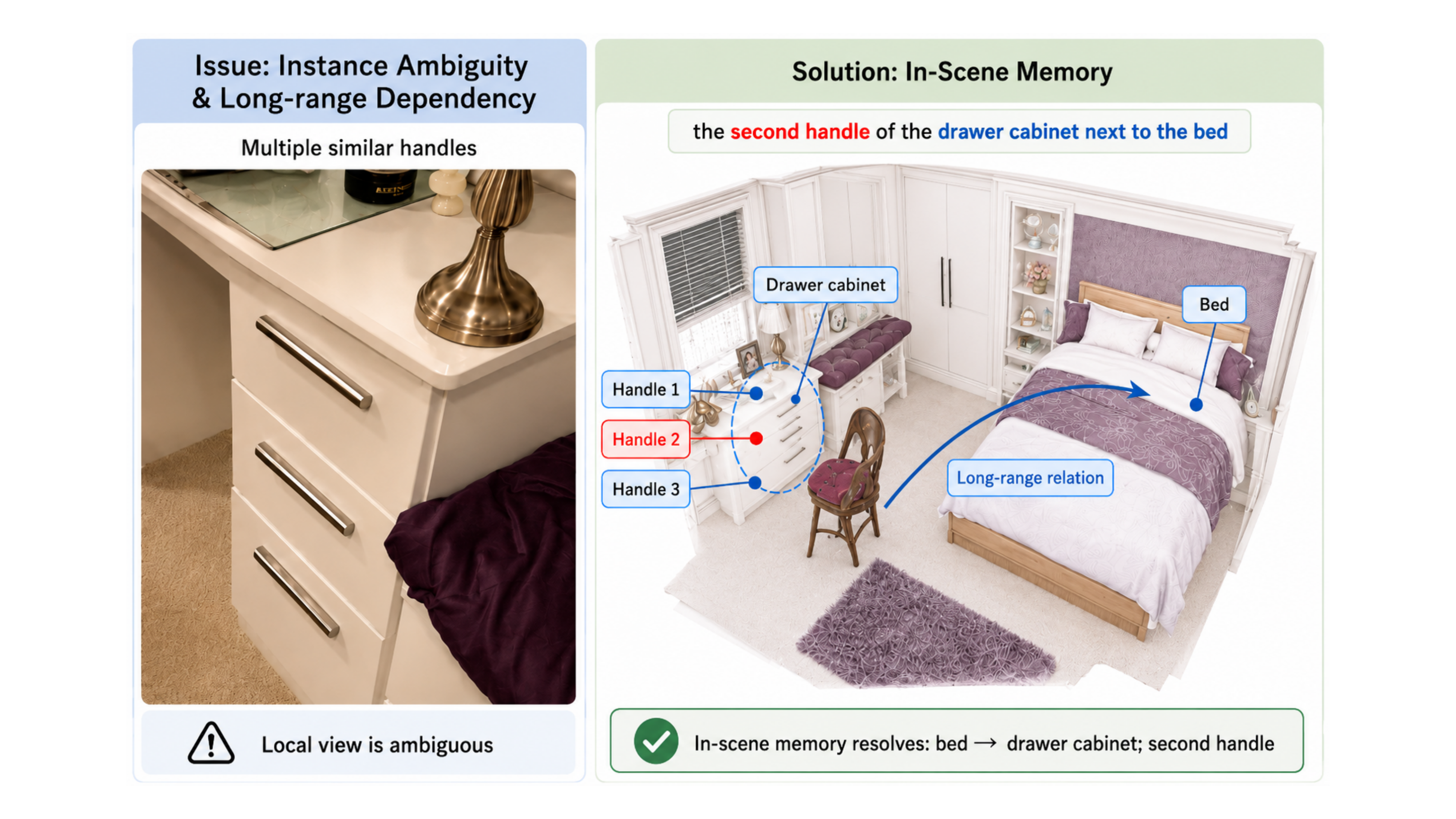}
        \label{fig:right}
    \end{subfigure}
\end{figure*}

Enabling a robot to ``close the second drawer from the top'' requires
far more than object recognition.
The agent must identify the exact 3D region that affords grasping
(not the drawer face, not the surrounding cabinet) and it must do so
for the \emph{correct} handle among potentially several that are
geometrically identical.
This problem, 3D functional affordance grounding, demands two
capabilities that existing vision-language pipelines handle poorly in
combination: fine-grained localization of small operable subregions,
and spatial disambiguation among multiple same-category instances.

The first difficulty is \emph{granularity ambiguity}.
Affordance regions are genuinely small: a door handle can occupy fewer
than $30\!\times\!30$ pixels in a $1920\!\times\!1440$ frame, and a
VLM prompted with ``drawer handle'' tends to return the entire drawer
panel rather than the graspable grip.
The second difficulty is \emph{instance disambiguation}.
Resolving a query like ``second drawer from the top'' requires knowing
the global spatial layout of all candidate handles in the scene, which
no single view can provide.
Moreover, the target instance may be spatially distant from the current
viewpoint or temporarily out of frame, making cross-view aggregation
over the full sequence essential.
Prior methods share a structural blind spot: they treat a posed RGB-D
sequence as a bag of independent 2D frames, discarding the 3D geometry
that ties those frames together and leaving both difficulties
unaddressed.

The key insight is that 3D geometry encodes exactly the signal that is
missing for both difficulties.
Humans resolve granularity ambiguity by memory: having operated many
cabinet handles, one recalls precisely which sub-region affords
grasping.
Once multi-view detection produces a set of 3D candidate instances,
their point-cloud positions encode a global spatial layout that supports
ordinal and relational reasoning.
We propose that \textsc{AffordMem} treats each difficulty as a distinct memory failure and addresses it with a dedicated geometric memory mechanism: \crossmem{} supplies the missing category-level spatial prior, and \inscmem{} supplies the missing global instance layout.

The first is \textbf{cross-scene affordance memory}.
The agent maintains a category-level memory bank of RGB images with
affordance regions rendered as mask overlays, built from previously
annotated scenes.
At query time, the most informative examples are recalled and provided
to the VLM as visual prompts, steering it toward the correct operable
subregion rather than the enclosing object.

The second is \textbf{in-scene spatial memory}.
As the agent processes the scene, it lifts 2D grounding results into a
3D candidate pool and organizes them into a structured scene graph,
recording each instance's 3D position and its spatial relations to
reference objects.
This structured memory enables the LLM to resolve spatial qualifiers
such as ``second from the top'' by reasoning over the global layout of
all candidates, including those that are distant or temporarily out of
frame.

Neither component requires model fine-tuning or target-scene annotation
at inference time; the cross-scene memory is built once from source-scene
annotations, while target scenes are processed using only the posed RGB-D
sequence available in a standard pipeline.
% ── CHANGE 2: filled in placeholder mAP numbers from Table 1 ────────────────
We evaluate on SceneFun3D~\citep{scenefun3d} and improve over the prior
training-free state of the art by $+3.23$ AP$_{50}$ on Split~0 and
$+3.7$ AP$_{50}$ on Split~1.
% ────────────────────────────────────────────────────────────────────────────
Ablation studies show complementary effects: \crossmem{} improves
fine-grained localization, while \inscmem{} provides the larger gain for
spatially qualified instance selection.

\paragraph{Contributions.}
\begin{itemize}
  \item We propose \crossmem{}: a category-level visual memory bank
    recalled at query time to guide a frozen VLM toward small operable
    subregions that text-only prompting consistently misses.
  \item We propose \inscmem{}: a structured in-scene spatial memory that
    accumulates 3D candidate instances and their spatial relations,
    enabling reliable selection under ordinal, relational, and
    long-distance spatial qualifiers.
\item Both components are training-free at inference time and require no
    target-scene annotation; together they set a new state of the art
    among training-free methods on SceneFun3D, with controlled ablations
    supporting independent gains from each.
\end{itemize}

% ─────────────────────────────────────────────────────────────────────────────
\section{Related Work}
\label{sec:related}

\paragraph{3D functional affordance localization.} The field of 3D affordance prediction has traditionally focused on isolated object point clouds, learning to predict interaction regions from geometry alone~\citep{3daffordnet,where2act}. SceneFun3D~\citep{scenefun3d} changed the problem setting substantially: 14.8K annotations across 230 full indoor scans, with free-form task descriptions, require methods to handle scene-level clutter, occlusion, and multi-instance disambiguation simultaneously. Fun3DU~\citep{fun3du} is the only prior method designed for this benchmark; it chains query parsing, per-frame VLM segmentation, and NeRF-based 3D lifting, but processes frames independently and targets whole-object masks rather than operable subregions. The concurrent TA3DAS~\citep{ta3das} achieves faster inference via CLIP-driven frame selection and Point Transformer refinement, again at the object level. The gap that neither system closes---and that we directly target---is the combination of subregion-level precision and multi-instance spatial disambiguation: knowing not just \emph{that} there is a handle, but \emph{exactly where} the grip surface lies, and \emph{which of several} such handles matches the spatial qualifier in the query.

\paragraph{Open-vocabulary 3D instance segmentation.} Two families of training-free methods have emerged for lifting 2D open-vocabulary predictions into 3D. Feature-lifting approaches~\citep{openscene,conceptfusion} encode CLIP features per 3D point and retrieve instances by text similarity; they produce object-level clusters that are too coarse for affordance localization. Mask-lifting approaches~\citep{openmask3d,openins3d,segment3d} run SAM~\citep{sam} or SAM2~\citep{sam2} frame-by-frame and fuse the resulting binary masks through multi-view voting, then score instances by aggregated CLIP similarity. Our 3D fusion stage belongs to this second family, but departs from it in a critical way: rather than feeding SAM a raw text prompt and scoring the output post-hoc, we inject a geometrically projected prior \emph{before} the segmentation stage, shifting the precision bottleneck upstream and replacing CLIP-based instance scoring with explicit spatial graph reasoning.

\paragraph{Language-guided 3D referring segmentation.} Supervised methods for 3D language grounding---ScanRefer~\citep{scanrefer}, Nr3D/Sr3D~\citep{nr3d}, and later works~\citep{3dres,mvggt}---train dedicated models to align text with 3D point-cloud features and require a pre-reconstructed scene at inference. In the training-free setting, the dominant strategy is to render 3D scenes into 2D images and query a VLM: Fun3DU uses per-frame video frames; AffordBot~\citep{affordbot} renders a $360^\circ$ panorama. The limitation shared by all photorealistic renders is that instances overlap and occlude one another, making it hard to resolve ordinal qualifiers such as ``second from the left'' when candidates are close together. We sidestep this by rendering a schematic top-down orthographic map that places every candidate at its correct 2D position without occlusion, trading photorealism for unambiguous spatial layout. % ── CHANGE 3: fixed dangling ref tab:ablation -> tab:adversarial_prompting ───
Table~\ref{tab:component_ablation} quantifies the benefit of this
choice directly.
% ────────────────────────────────────────────────────────────────────────────

\paragraph{Visual prompting and geometry-conditioned priors.} Visual prompting~\citep{visualprompt} adapts frozen models by augmenting their input images with task-relevant visual context. SegGPT~\citep{seggpt} uses reference image--mask pairs to condition segmentation on unseen categories; MIFAG~\citep{mifag} constructs affordance priors by aggregating regions from human-object interaction images. Both rely on 2D image retrieval, which introduces two independent error sources: an appearance gap when the retrieved image depicts a different furniture style, and a viewpoint gap when the retrieved image was captured from a different angle. \crossmem{} eliminates the viewpoint gap by construction---the projection is computed from the exact camera pose of the query frame---and reduces the appearance gap by aggregating a category-level distribution over many scenes rather than relying on a single retrieved exemplar.

\begin{figure}[t]
  \centering
    \caption{\textbf{\textsc{AffordMem} overview.}
    \textbf{(A)} A frozen VLM parses the query into a structured triple
    (\texttt{CTX}, \texttt{INT}, spatial qualifier).
    \textbf{(B)} \crossmem{} uses retrieved category-level masked images
    as a geometric prior to guide fine-grained subregion grounding, fusing
    multi-view masks into 3D candidates.
    \textbf{(C)} \inscmem{} organizes candidates into a global scene map
    and resolves the spatial qualifier via LLM reasoning.}
  \includegraphics[width=\textwidth]{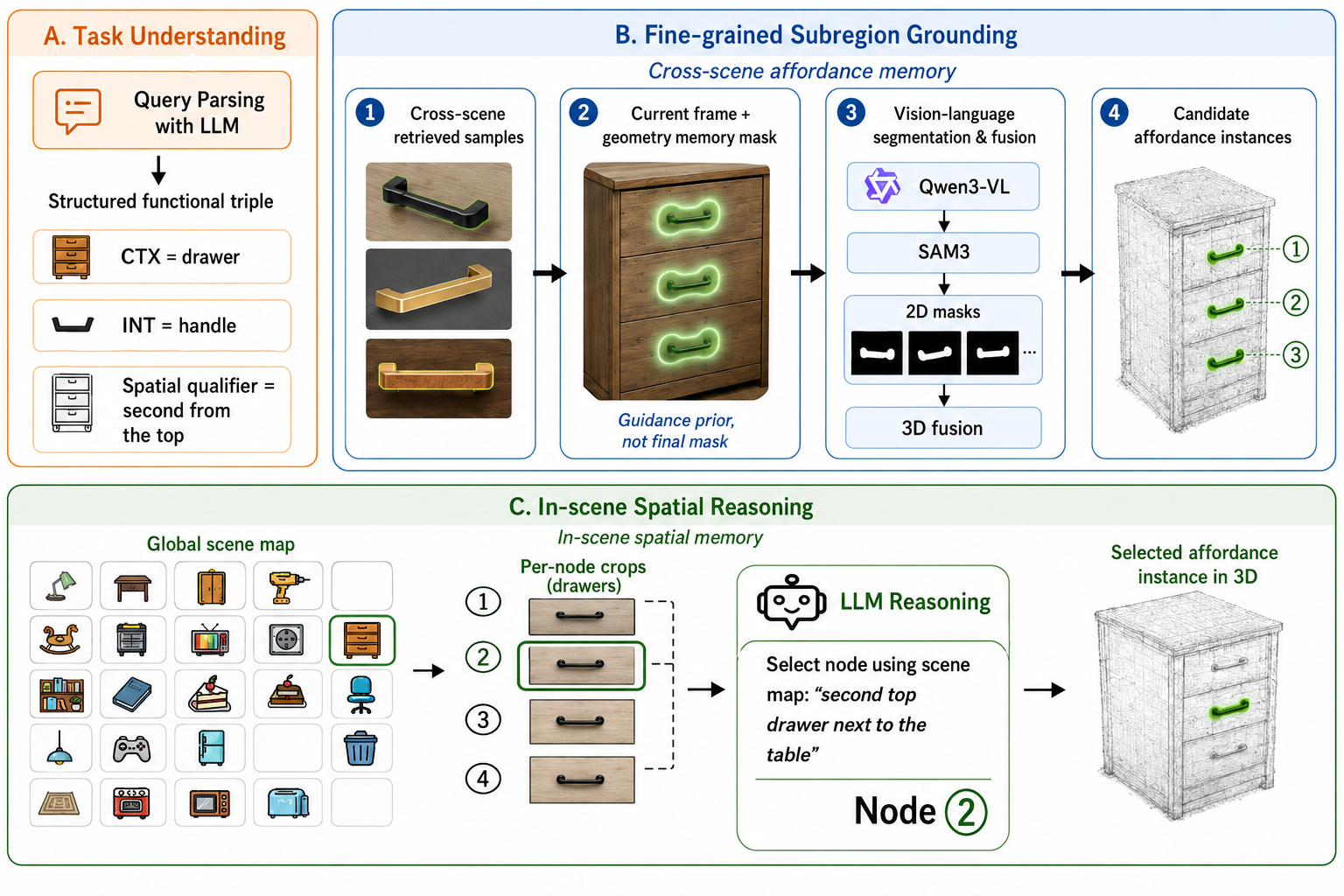}
  \label{fig:overview}
\end{figure}

% ─────────────────────────────────────────────────────────────────────────────
\section{Method}
\label{sec:method}

% ─────────────────────────────────────────────────────────────────────────────

\paragraph{Problem statement.}
We address \emph{functional affordance grounding} in 3D scenes.
The input is a posed RGB-D sequence
$\mathcal{F} = \{(I_t,\, D_t,\, K_t,\, T_t)\}_{t=1}^{T}$,
where $I_t \in \mathbb{R}^{H \times W \times 3}$ is an RGB frame,
$D_t \in \mathbb{R}^{H \times W}$ the corresponding depth map,
$K_t$ the camera intrinsic matrix, and $T_t \in \mathrm{SE}(3)$ the
camera-to-world pose.
Together, these frames densify into a scene point cloud
$P \subset \mathbb{R}^3$.
A natural-language functional query $q$ (e.g.,
\textit{``the handle of the second drawer from the top''}) specifies
an \emph{interactive element} (a small, operable sub-region of a larger
object).
The goal is to predict a 3D instance mask $\hat{M} \subseteq P$ that
tightly covers that element and no other.

\paragraph{Method overview.}
Given query $q$ and sequence $\mathcal{F}$, \textsc{AffordMem} proceeds
in three stages.
First, $q$ is parsed into an object label, an interaction label, and a
spatial descriptor (Section~\ref{sec:parsing}).
Second, \crossmem{} (Section~\ref{sec:memory}) leverages a cross-scene
geometric prior to guide the VLM toward the operable sub-region, and fuses the resulting 2D masks into a 3D
candidate pool.
Third, \inscmem{} (Section~\ref{sec:inscene}) organizes the candidates
into a structured scene graph and selects the instance matching the
spatial descriptor via language-guided reasoning. The overview pipeline is shown in Fig. \ref{fig:overview}.

\subsection{Functional Query Decomposition}
\label{sec:parsing}

A functional query $q$ carries two distinct types of information:
\emph{what to look for} (the object and its operable part) and
\emph{which one to pick} (a spatial qualifier disambiguating among
identical instances).
We parse $q$ into a structured triple
$(\ell_{\text{ctx}},\, \ell_{\text{int}},\, \sigma)$,
where $\ell_{\text{ctx}}$ is the reference object label
(e.g., ``drawer''), $\ell_{\text{int}}$ the interactive element label
(e.g., ``handle''), and $\sigma$ the spatial relational descriptor
(e.g., ``second from the top'').
The labels $\ell_{\text{ctx}}$ and $\ell_{\text{int}}$ are used to
construct text prompts for SAM3, which generates initial candidate masks
for the reference object and interactive region respectively, bounding
the search space for subsequent grounding.
The spatial descriptor $\sigma$ is forwarded to the instance selection
stage (\inscmem{}, Section~\ref{sec:inscene}).

\subsection{\Crossmem{}}
\label{sec:memory}

When localizing a functional element such as a ``door handle,'' a VLM
faces an inherent granularity ambiguity: should it segment the entire
door, the handle assembly, or only the directly graspable grip?
Humans resolve this by recalling past interactions: having opened many
doors, one knows precisely which sub-region to grasp.
\crossmem{} instantiates this intuition as a category-level affordance-region prior: unlike few-shot visual prompting, which retrieves appearance-similar images, \crossmem{} recalls \emph{where} affordances of a category geometrically tend to appear within their parent objects, providing a viewpoint-consistent spatial prior rather than a visual exemplar.

\paragraph{Memory construction.}
The memory is organized by semantic category $c$, where
$\mathcal{G}_c \subset \mathbb{R}^3$ denotes the set of 3D points
annotated as category $c$ in the ground-truth annotations of previously
seen scenes.
We first identify frames in which the affordance region is visible by
checking whether any point in $\mathcal{G}_c$ projects within the image
bounds with a valid depth reading.
For each such frame $t$, we project $\mathcal{G}_c$ into the image
plane using pose $T_t$ and intrinsics $K_t$:
\begin{equation}
  (u_i,\, v_i) = \pi\!\bigl(K_t \cdot T_t^{-1} \cdot p_i\bigr),
  \qquad p_i \in \mathcal{G}_c.
  \label{eq:proj}
\end{equation}
Raw projection suffers from false positives at occlusion boundaries,
which we suppress with a \emph{depth-consistency filter}.
The depth residual for each projected point is
\begin{equation}
  r_i = z_i^{\mathrm{proj}} - z_t^{\mathrm{sensor}}(u_i, v_i),
  \label{eq:residual}
\end{equation}
and a point is retained when
\begin{equation}
  |r_i - \hat{\mu}_t| < \tau_t,
  \qquad
  \tau_t = \max\!\bigl(\tau_{\min},\; k \cdot \mathrm{MAD}_t\bigr),
  \label{eq:depth}
\end{equation}
where $\hat{\mu}_t$ and $\mathrm{MAD}_t$ are the per-frame median and
median absolute deviation of $\{r_i\}$; the MAD-based adaptive threshold
avoids manual tuning across scenes.
The filtered projection yields a binary mask for each frame, from which
we compute a quality score
\begin{equation}
  s_t = w_1 \cdot \mathrm{centrality}_t + w_2 \cdot \mathrm{proximity}_t,
  \label{eq:score}
\end{equation}
where $\mathrm{centrality}_t$ measures how close the mask centroid is
to the image center and $\mathrm{proximity}_t$ is the inverse
camera-to-object distance (both normalized per sequence).
The top-$K$ frames are selected globally across all scenes in the
memory bank for category $c$, retaining the most informative views
regardless of which scene they originate from.
For each selected frame, the binary mask is filled with convex-hull
for spatial coherence, rendered as a colored overlay onto $I_t$ to
produce a masked image $\hat{I}_t^{(c)}$.
The memory bank for category $c$ is the collection
$\{\hat{I}_t^{(c)}\}_{t \in \mathrm{top}\text{-}K}$.

\paragraph{Memory recall and VLM prompting.}
At query time, the interaction label $\ell_{\mathrm{int}}$ is matched
to the memory bank by exact string matching against the category index
$c$.
The corresponding masked images $\{\hat{I}_t^{(c)}\}$ are provided to
the VLM as visual examples alongside the query frame, with the prompt:
\begin{tcolorbox}[
  colback=gray!8, colframe=gray!40, boxrule=0.4pt, arc=2pt,
  left=6pt, right=6pt, top=4pt, bottom=4pt, fontupper=\small
]
The highlighted regions show where \textit{\{INT label\}} instances
appear in similar scenes. Locate all \textit{\{INT label\}} instances
in this image and return tight bounding boxes.
\end{tcolorbox}
To further sharpen granularity, the VLM is asked to produce paired
proposals: a positive box for the directly operable core and a negative
box for visually similar but non-operable regions (e.g., the mounting
base of a handle).
Only positive boxes are passed back to SAM3 to produce fine-grained 2D
segmentation masks, which are subsequently fused into 3D
(Section~\ref{sec:inscene}).

\subsection{\Inscmem{}}
\label{sec:inscene}

Functional scenes routinely contain multiple identical instances of the
same interactive element, such as three drawer handles on a cabinet or
a row of light switches, and the spatial descriptor $\sigma$ is the
only signal that distinguishes them.
Moreover, the target instance may be spatially distant from the camera
or unobserved in any single frame, making cross-view aggregation
essential.
Resolving $\sigma$ therefore requires global scene context that no
single view or isolated crop can provide.
\inscmem{} is not a general-purpose scene graph: it is a spatial memory purpose-built for resolving linguistic qualifiers over geometrically identical affordance candidates, organizing instances by their 3D positions and part-object relationships rather than semantic attributes.

\paragraph{3D candidate pool construction.}
The 2D masks produced by SAM3 are backprojected into $P$ using
depth-consistent projection.
For each point $p_i \in P$, we accumulate its visibility count
$n_i^{\mathrm{vis}}$ and foreground support count $n_i^{\mathrm{fg}}$
across all frames, and assign it to a foreground mask when
\begin{equation}
  \frac{n_i^{\mathrm{fg}}}{n_i^{\mathrm{vis}}} > \rho_0
  \quad\text{and}\quad
  n_i^{\mathrm{vis}} > \theta_{\mathrm{vis}},
  \label{eq:voting}
\end{equation}
where $\rho_0$ is a Wilson lower-bound threshold and
$\theta_{\mathrm{vis}}$ a minimum visibility requirement.
Overlapping hypotheses are merged with DBSCAN clustering, yielding a
candidate pool $\{\hat{M}^{(j)}\}$.

\paragraph{Scene graph construction and instance selection.}
Each candidate becomes a node annotated with its 3D centroid, AABB, and
semantic type (\texttt{CTX} for reference objects, \texttt{INT} for
interactive regions), with affordance nodes linked to their parent
\texttt{CTX} nodes to encode object--part relationships.
The graph is serialized as JSON and provided to the LLM alongside a
top-down orthographic render of the point cloud with AABB boxes labeled
by node ID, and a per-node RGB crop for appearance confirmation
(full prompt template in Appendix~\ref{app:prompts}).
The LLM returns the node ID of the matching instance; the corresponding
mask $\hat{M}^{(j)}$ is the final prediction.

% ─────────────────────────────────────────────────────────────────────────────

% ─────────────────────────────────────────────────────────────────────────────
% ─────────────────────────────────────────────────────────────────────────────
\section{Experiments}
\label{sec:experiments}
% ─────────────────────────────────────────────────────────────────────────────

\subsection{Experimental Setup}
\label{sec:setup}

\paragraph{Dataset.}
We evaluate on \textbf{SceneFun3D}~\citep{scenefun3d}, the only
large-scale benchmark for 3D functional affordance segmentation.
The dataset comprises 230 real-world indoor scenes, each with an
average of 1800 high-resolution RGB-D frames and approximately 15
natural-language task descriptions per scene.
Following Fun3DU~\citep{fun3du}, we report results on two protocols:
\textbf{Split~0} (30 scenes, validation set) and \textbf{Split~1}
(200 scenes, training set).

\paragraph{Metrics.}
Following the SceneFun3D benchmark~\citep{scenefun3d}, we report
Average Precision at IoU thresholds of 0.25 and 0.50 (\textbf{AP25},
\textbf{AP50}).
We additionally report mean IoU (\textbf{mIoU}) and Average Recall
at the same thresholds (\textbf{AR25}, \textbf{AR50}).

\paragraph{Implementation details.}
All language-based stages of \textsc{AffordMem} use the same Qwen3-VL-32B~\citep{qwenvl} model for query parsing, VLM grounding, and instance selection, ensuring that performance differences arise from the proposed memory mechanisms rather than model ensembling.
Segmentation uses SAM3~\citep{sam3} on a single NVIDIA A100 80\,GB GPU.
Depth consistency filter: $k{=}3$, $\tau_{\min}{=}0.05$\,m.
Multi-view voting: $\rho_0{=}0.70$, $\theta_{\mathrm{vis}}{=}3$,
DBSCAN $\varepsilon{=}0.03$\,m, merge IoU threshold
$\theta_{\mathrm{IoU}}{=}0.30$, recall threshold
$\theta_{\mathrm{rec}}{=}0.60$.
Frame selection: $K{=}20$, $w_1{=}w_2{=}0.5$.
All hyperparameters are fixed across all scenes and splits without
scene-specific tuning.
For cross-split evaluation, the memory bank for Split~0 is constructed
from Split~1 scenes and vice versa, ensuring that no test-scene
geometry appears in the memory at evaluation time.
Thus, the memory bank uses only source-scene annotations and never
accesses target-scene geometry or target-scene affordance labels during
evaluation.

\paragraph{Baselines.}
We compare against three categories of methods.
Open-vocabulary 3D segmentation methods include
\textbf{OpenMask3D}~\citep{openmask3d},
\textbf{LERF}~\citep{lerf}, and
\textbf{OpenIns3D}~\citep{openins3d}.
Training-based functionality segmentation methods include
\textbf{TASA}~\citep{ta3das} and
\textbf{AffordBot}~\citep{affordbot}, both trained on Split~1
and evaluated on Split~0 only.
Among training-free methods, we compare against
\textbf{Fun3DU}~\citep{fun3du} , which is the
only prior method evaluated on both splits.

% ── Main results table ───────────────────────────────────────────────────────
\begin{table*}[t]
    \centering
    \small
    \caption{Quantitative comparison on SceneFun3D. All numbers are
    obtained with the official evaluation script; results for prior
    methods are taken from their respective papers and verified against
    the script output. Missing entries for TASA reflect values not
    reported in the original paper. Training-based methods are excluded
    from Split~1 evaluation as that split is used for their training.
    For our method, the Split~0 memory bank is built from Split~1 scenes
    and vice versa, ensuring no test-scene overlap.}
    \label{tab:main_results}
    \setlength{\tabcolsep}{4pt}
    \resizebox{\textwidth}{!}{%
    \begin{tabular}{lccccc|ccccc}
        \toprule
        & \multicolumn{5}{c|}{\textbf{Split~0} (30 scenes)}
        & \multicolumn{5}{c}{\textbf{Split~1} (200 scenes)} \\
        \cmidrule(lr){2-6}\cmidrule(lr){7-11}
        Method
          & AP$_{50}$ & AP$_{25}$ & AR$_{50}$ & AR$_{25}$ & mIoU
          & AP$_{50}$ & AP$_{25}$ & AR$_{50}$ & AR$_{25}$ & mIoU \\
        \midrule
        \multicolumn{11}{l}{\emph{Open-vocabulary 3D segmentation methods}} \\
        OpenMask3D~\citep{openmask3d}
          & 0.2  & 0.4  & 24.5 & 27.0 & 0.2
          & 0.0  & 0.0  &  1.4 &  2.6 & 0.1 \\
        LERF~\citep{lerf}
          & 0.0  & 0.0  & 35.1 & 36.0 & 0.0
          & 0.0  & 0.0  & 24.6 & 25.1 & 0.0 \\
        OpenIns3D~\citep{openins3d}
          & 0.0  & 0.0  & 46.7 & 51.5 & 0.1
          & 0.0  & 0.0  & 37.1 & 39.9 & 0.1 \\
        \midrule
        \multicolumn{11}{l}{\emph{Training-based 3D functionality segmentation methods}} \\
        TASA~\citep{ta3das}
          & 26.9 & 28.6 & --   & --   & 19.7
          & --   & --   & --   & --   & --   \\
        AffordBot~\citep{affordbot}
          & 20.91 & 24.76 & 18.99 & 22.84 & 14.42
          & --   & --   & --   & --   & --   \\
        \midrule
        \multicolumn{11}{l}{\emph{Training-free 3D functionality segmentation methods}} \\
        Fun3DU~\citep{fun3du}
          & 16.9  & 33.3  & 38.2  & 46.7  & 15.2
          & 12.6  & 23.1  & 32.9  & 40.5  & 11.5 \\
        \textsc{AffordMem}
          & \textbf{20.13} & \textbf{41.66}
          & \textbf{44.04} & \textbf{50.62} & \textbf{17.32}
          & \textbf{16.3}  & \textbf{32.7}
          & \textbf{41.4}  & \textbf{47.8}  & \textbf{14.0}  \\
        \bottomrule
    \end{tabular}}
\end{table*}
% ─────────────────────────────────────────────────────────────────────────────

% ── Component ablation table ─────────────────────────────────────────────────
\begin{table}[ht]
    \centering
    \small
    \setlength{\tabcolsep}{8pt}
    \caption{\textbf{Component ablation.}
    We ablate adversarial prompting (\textsc{no ap}) and the scene graph
    (\textsc{no sg}), and include a Fun3DU-style diagnostic with Qwen3-VL
    and SAM3 as a lower-bound reference.}
    \label{tab:component_ablation}
    \begin{tabular}{lrrrrr}
        \toprule
        Method & AP$_{50}$ & AP$_{25}$ & AR$_{50}$ & AR$_{25}$ & mIoU \\
        \midrule
        Fun3DU-style (Qwen3+SAM3)
          &  7.14 & 17.86 & 14.29 & 17.86 &  7.06 \\
        \midrule
        \textsc{AffordMem} w/o AP
          & 19.67 & 37.21 & \textbf{38.02} & 42.61 & 15.58 \\
        \textsc{AffordMem} w/o SG
          & 13.21 & 29.63 & 32.88 & 42.53 & 12.78 \\
        \textsc{AffordMem}
          & \textbf{20.39} & \textbf{40.13} & 37.19
          & \textbf{45.82} & \textbf{16.33} \\
        \bottomrule
    \end{tabular}
\end{table}
% ─────────────────────────────────────────────────────────────────────────────

% ── Memory component ablation table ──────────────────────────────────────────
\begin{table}[ht]
    \centering
    \small
    \setlength{\tabcolsep}{8pt}
\caption{\textbf{Effect of memory-guided grounding.}
We compare direct SAM3 text prompting with our memory-guided grounding
pipeline (MG-Grounding), where cross-scene affordance memory guides the VLM before SAM3
mask refinement. Check marks indicate enabled components.}
    \label{tab:memory_ablation}
    \begin{tabular}{cc|rrrrr}
        \toprule
        SAM3 & MG-Grounding
          & AP$_{50}$ & AP$_{25}$ & AR$_{50}$ & AR$_{25}$ & mIoU \\
        \midrule
        \checkmark &  
          & 14.80 & 30.45 & 34.04 & 41.81 & 13.96 \\
        \checkmark & \checkmark 
          & \textbf{20.39} & \textbf{40.13} & 37.19
          & \textbf{45.82} & \textbf{16.33} \\
        \bottomrule
    \end{tabular}
\end{table}
% ─────────────────────────────────────────────────────────────────────────────

% ── Memory size ablation table ────────────────────────────────────────────────
\begin{table}[ht]
    \centering
    \small
    \setlength{\tabcolsep}{8pt}
    \caption{Effect of memory bank size $K$ on the representative Split~0
    ablation subset.
    The full system uses $K{=}20$ for robustness across categories with
    fewer high-quality views.}
    \label{tab:memory_size}
    \begin{tabular}{lrrrrr}
        \toprule
        $K$ & AP$_{50}$ & AP$_{25}$ & AR$_{50}$ & AR$_{25}$ & mIoU \\
        \midrule
         5 & 20.08 & 38.59 & \textbf{38.54} & 43.42 & 16.12 \\
        20 & \textbf{20.39} & \textbf{40.13} & 37.19 & \textbf{45.82} & \textbf{16.33} \\
        30 & 20.17 & 39.71 & 37.63 & 45.31 & 16.09 \\
        \bottomrule
    \end{tabular}
\end{table}
% ─────────────────────────────────────────────────────────────────────────────

\subsection{Main Results}
\label{sec:main}

Table~\ref{tab:main_results} reports quantitative results on
SceneFun3D across all methods.
\textsc{AffordMem} achieves state-of-the-art performance among
training-free methods on both evaluation splits, outperforming Fun3DU
by $+3.23$ AP$_{50}$ and $+8.36$ AP$_{25}$ on Split~0, and by
$+3.7$ AP$_{50}$ and $+9.6$ AP$_{25}$ on Split~1.
These gains are consistent across all reported metrics, with
improvements in AR$_{50}$, AR$_{25}$, and mIoU on both splits.

Compared to training-based methods evaluated on Split~0,
\textsc{AffordMem} surpasses AffordBot by a substantial margin on
AP$_{25}$ ($24.76 \to 41.66$), indicating substantially tighter
localization of operable subregions despite requiring no
scene-specific training or supervision.
\textsc{AffordMem} also exceeds TASA on AP$_{25}$ ($28.6 \to 41.66$)
while remaining competitive on AP$_{50}$ ($26.9$ vs.\ $20.13$),
a gap we attribute to the stricter IoU requirement at 0.50 being more
sensitive to the bounding precision of our 3D fusion stage.

Open-vocabulary 3D segmentation methods (OpenMask3D, LERF, OpenIns3D)
achieve near-zero AP across both thresholds, confirming that
general-purpose open-vocabulary segmentation is fundamentally
ill-suited to the subregion-level precision required by functional
affordance grounding.
Their relatively higher AR values suggest these methods do produce
relevant regions among a large set of proposals, but fail to rank the
correct instance highly enough for AP to reflect meaningful performance.

\subsection{Ablation Study}
\label{sec:ablation}

Ablations are conducted on a representative subset of Split~0,
stratified to span strong, average, and failure cases of the full
system, ensuring representative coverage across scene types and query
difficulty.
Results are shown in Tables~\ref{tab:component_ablation},
\ref{tab:memory_ablation}, and~\ref{tab:memory_size}.

\paragraph{Effect of the scene graph (\inscmem{}).}
Removing the structured scene graph and replacing instance selection
with flat per-frame grounding (\textsc{AffordMem} w/o SG in
Table~\ref{tab:component_ablation}) causes a drop of $7.18$
AP$_{50}$, $10.50$ AP$_{25}$, and $3.55$ mIoU relative to the full
system.
This is the largest single-component degradation, supporting the
importance of explicit 3D spatial reasoning over the candidate pool
for resolving ordinal and relational qualifiers.
The drop is especially pronounced on AP$_{50}$, consistent with the
hypothesis that incorrect instance selection returns a spatially
plausible but geometrically wrong mask, yielding low overlap with the
ground truth.

\paragraph{Effect of adversarial prompting (\crossmem{}).}
Removing the positive/negative box pairing (\textsc{AffordMem} w/o AP
in Table~\ref{tab:component_ablation}) reduces AP$_{25}$ by $2.92$
points and mIoU by $0.75$ points.
The consistent drop across all metrics indicates that adversarial
prompting provides a targeted improvement to fine-grained localization:
by explicitly requesting a negative box for visually similar but
non-operable regions, the VLM is steered away from returning loose
bounding boxes that encompass the entire parent object.

\paragraph{Effect of cross-scene memory.}
Table~\ref{tab:memory_ablation} compares direct SAM3 text prompting with
our memory-guided grounding pipeline. Direct SAM3 prompting provides a
simple baseline that produces 2D masks from the interaction label alone.
Adding cross-scene affordance memory guides the grounding process with
category-level visual exemplars before SAM3 mask refinement, improving
performance by $5.59$ AP$_{50}$, $9.68$ AP$_{25}$, and $2.37$ mIoU.
These gains suggest that memory-guided grounding provides a stronger
fine-grained localization signal than directly prompting SAM3 with text
alone.

\paragraph{Effect of memory bank size.}
Table~\ref{tab:memory_size} compares $K \in \{5, 20, 30\}$ recalled
frames.
$K{=}20$ achieves the best AP$_{50}$, AP$_{25}$, AR$_{25}$, and mIoU,
while $K{=}5$ retains the highest AR$_{50}$.
We attribute the slight AP degradation at $K{=}30$ to context length
saturation: additional masked images introduce redundant views that
dilute rather than reinforce the spatial prior within the VLM's context
window.
The full system therefore uses $K{=}20$ as a setting that balances
peak performance with robustness across categories that have fewer
high-quality views in the memory bank.

\paragraph{Diagnostic comparison to a Fun3DU-style pipeline.}
We additionally evaluate a Fun3DU-style diagnostic by replacing the
original VLM and segmentation components with Qwen3-VL and SAM3, while
keeping the pipeline frame-local and excluding our memory modules.
This diagnostic obtains $7.14$ AP$_{50}$ and $17.86$ AP$_{25}$,
below both the official Fun3DU result and our full system.
Since Fun3DU is optimized around its original multi-stage design,
including frame selection, prompting, segmentation, and 3D lifting, we
treat this reimplementation only as a diagnostic rather than a primary
baseline. The result indicates that simply replacing backbones is
insufficient without addressing subregion grounding and global instance
disambiguation.

\subsection{Qualitative Results}
\label{sec:qual}

\begin{figure}[t]
  \centering
  \includegraphics[width=\linewidth]{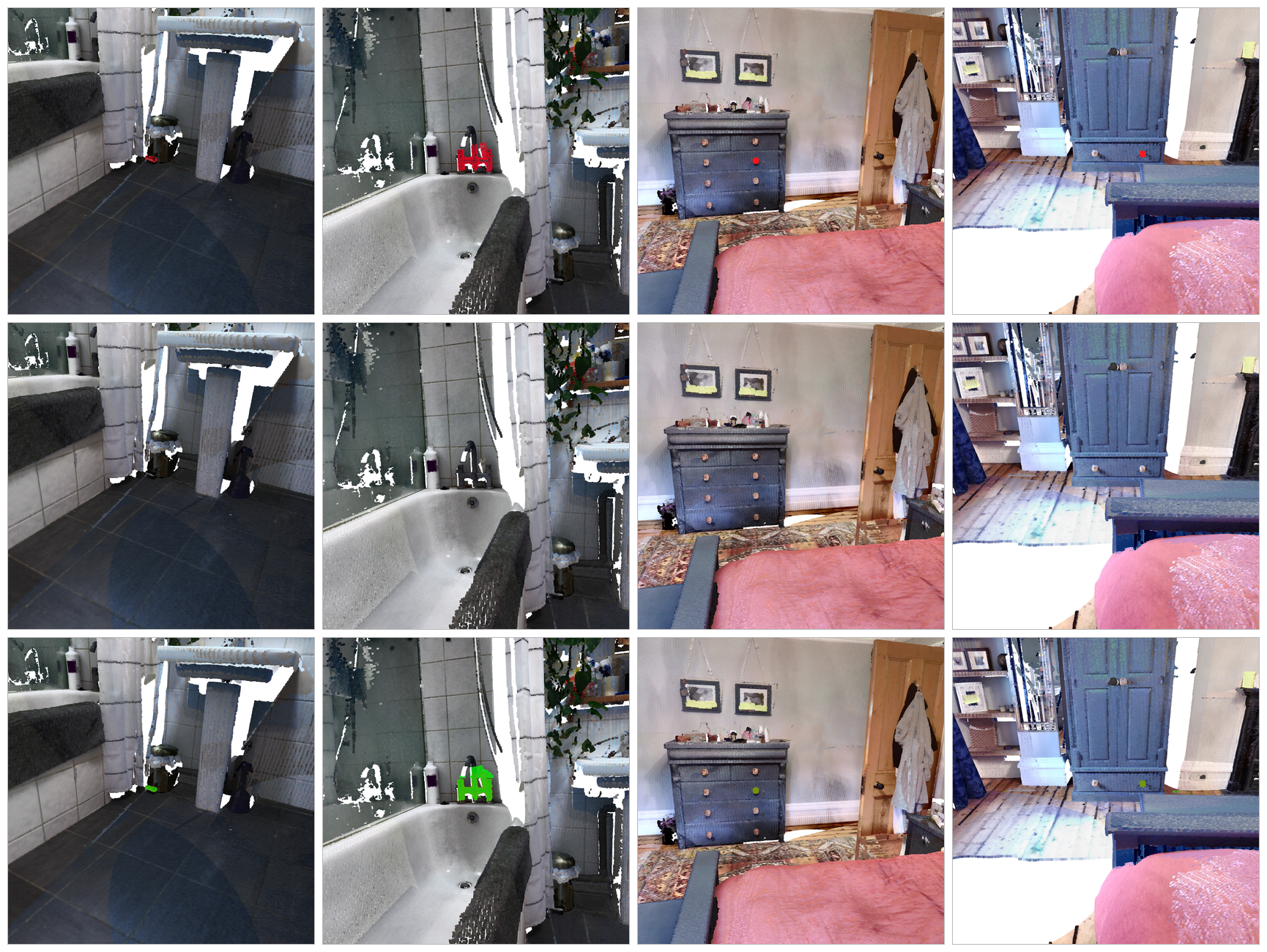}
\caption{\textbf{Qualitative comparison on four representative cases.}
Each column shows one affordance grounding example.
The first row shows the ground-truth target regions, the second row shows
Fun3DU failure cases where no corresponding target mask is produced, and
the third row shows predictions from \textsc{AffordMem}.
\textsc{AffordMem} recovers compact affordance regions for small and
visually ambiguous targets that are missed by the frame-local baseline.}
\label{fig:qual}
\end{figure}

Figure~\ref{fig:qual} shows qualitative comparisons across four
representative affordance grounding cases. The first row shows the
ground-truth target regions, while the second row shows Fun3DU failure
cases where no corresponding target mask is produced. These examples
highlight a common failure mode of frame-local grounding pipelines:
small operable parts can be missed when they occupy only a small image
region, appear in cluttered surroundings, or are visually similar to
nearby non-target structures.

In contrast, \textsc{AffordMem} recovers compact affordance regions in
the third row. The cross-scene memory provides category-level visual
guidance for where operable parts such as handles or knobs usually
appear, helping the VLM focus on the actionable subregion rather than
the surrounding object. For scenes with repeated candidates, the
in-scene spatial memory further supports consistent target selection
using the global scene layout. These qualitative results are consistent
with the quantitative gains in Tables~\ref{tab:memory_ablation}
and~\ref{tab:component_ablation}.

% ─────────────────────────────────────────────────────────────────────────────
\section{Conclusion}
\label{sec:conclusion}
% ─────────────────────────────────────────────────────────────────────────────

% ── CHANGE 10: added conclusion body (was entirely empty) ────────────────────

We presented \textsc{AffordMem}, a training-free framework for 3D functional affordance grounding that addresses two complementary failure modes of existing vision-language pipelines. \crossmem{} provides a category-level geometric prior that steers a frozen VLM toward small operable subregions rather than enclosing objects, while \inscmem{} organizes 3D candidate instances into a structured scene graph to support spatial reasoning over ordinal and relational qualifiers. Together, the two components achieve state-of-the-art performance among training-free methods on SceneFun3D, with ablations supporting their complementary contributions.

\paragraph{Limitations \& Future directions.} The current memory bank relies on category-level string matching between the parsed interaction label and the memory index; categories not seen during memory construction therefore receive no prior. Future work could replace exact matching with embedding-based retrieval to generalize to unseen or semantically related affordance categories. The scene graph currently encodes only geometric relations; integrating functional, material, or state attributes could further improve disambiguation in scenes with highly similar candidate instances. Finally, extending \textsc{AffordMem} to dynamic scenes, where objects may move across frames, remains an important direction for future work.

% \begin{ack}
% Funding information omitted for anonymity.
% \end{ack}

\clearpage

\bibliographystyle{plain}
\bibliography{references}

% ─────────────────────────────────────────────────────────────────────────────
\appendix
% ─────────────────────────────────────────────────────────────────────────────

\section{Memory Mask: Category-Level Prior vs.\ Instance Ground Truth}
\label{app:memory_clarification}

A key concern is whether the \gemmask{} constitutes label leakage.
It does not, for three reasons.
First, $\mathcal{G}_c$ is constructed exclusively from \emph{training}
scenes; the test scene's ground truth is never accessed.
Second, $\mathcal{G}_c$ is a category-level aggregate projected from
\emph{different} scenes' point clouds; because furniture geometry, scale,
and pose vary across scenes, the projection will not precisely align with
the test instance---it provides a spatial hint, not a direct answer.
% ── CHANGE 12: fixed dangling ref tab:ablation -> tab:adversarial_prompting ──
Third, the memory ablation (Table~\ref{tab:memory_ablation})
shows that VLM grounding \emph{without} any memory mask already
substantially outperforms non-VLM baselines, indicating that the
\gemmask{} is a prompt improvement, not a shortcut.
% ────────────────────────────────────────────────────────────────────────────

\section{Prompt Templates}
\label{app:prompts}

\paragraph{Stage~1: query parsing.}
{\small
\begin{verbatim}
You translate a natural-language interaction description 
into structured grounding for a robotic manipulator.

Given a description, identify:
1. contextual_object: the concrete physical object that must be operated on first. 
It must be explicitly mentioned; use "None" if unavailable.
2. interactive_objects: concrete physical interfaces involved in the task
such as handles, knobs, switches, remotes, sockets, ports, or keyholes.
3. functional_object_candidates: ranked intrinsic functional parts 
of the contextual_object relevant to the task.
4. action: choose exactly one from 
[rotate, key_press, tip_push, hook_pull, pinch_pull, hook_turn, foot_push, plug_in, unplug].
5. spatial_relation [X, Y]: extract at most one explicit spatial relation 
if mentioned using [contextual_object, referenced_object]; otherwise output "N/A".

Classify the physical execution implicitly:
- intrinsic manipulation: directly operate a part of the target object.
- external-mediated manipulation: operate a separate control/tool to change
the target object state.
- insertion/receptacle manipulation: insert/remove an object into/from an interface.

Output ONLY the following fields in this exact order:

contextual_object
interactive_objects
......
action
spatial_relation [X, Y] or N/A
original_prompt

Preserve original_prompt exactly. Do not output reasoning or explanations.
\end{verbatim}
}

\paragraph{Stage~2: Qwen Grounding(adversarial prompting ).}
{\small
\begin{verbatim}
You are helping to localize affordance parts for a given instruction.

Instruction: "{instruction}"
Affordance category: "{category}"

For each relevant visible instance, return exactly two tight bounding boxes:
part_index 1: 
the minimum directly operable region needed for the action
such as the part to touch, press or manipulate.
part_index 0: the connected non-core part of the same object
such as base, mount, panel, housing, body, stem, or support.

The two boxes must belong to the same object instance.
Exclude unrelated objects, background, surrounding surfaces
black masked regions and neighboring instances.

Return JSON only as a list of objects with fields:
bbox_2d and part_index."""
\end{verbatim}
}

\paragraph{Stage~3: instance selection.}
{\small
\begin{verbatim}
You are helping to select the most appropriate affordance node
from a scene graph based on a given instruction.

Instruction: "{instruction}"

Available affordance nodes

Please analyze the instruction and select the most 
appropriate node that would be needed to perform this action. Consider:
1. The action verb in the instruction
2. The object being acted upon
3. The semantic meaning and context
Respond with ONLY the number (1-{len(affordance_nodes)}) 
of the most appropriate node. 
Selected node number:
\end{verbatim}
}

\newpage
\section*{NeurIPS Paper Checklist}

%%% BEGIN INSTRUCTIONS %%%
The checklist is designed to encourage best practices for responsible machine learning research, addressing issues of reproducibility, transparency, research ethics, and societal impact. Do not remove the checklist: {\bf The papers not including the checklist will be desk rejected.} The checklist should follow the references and follow the (optional) supplemental material.  The checklist does NOT count towards the page
limit. 

Please read the checklist guidelines carefully for information on how to answer these questions. For each question in the checklist:
\begin{itemize}
    \item You should answer \answerYes{}, \answerNo{}, or \answerNA{}.
    \item \answerNA{} means either that the question is Not Applicable for that particular paper or the relevant information is Not Available.
    \item Please provide a short (1--2 sentence) justification right after your answer (even for \answerNA). 
   % \item {\bf The papers not including the checklist will be desk rejected.}
\end{itemize}

{\bf The checklist answers are an integral part of your paper submission.} They are visible to the reviewers, area chairs, senior area chairs, and ethics reviewers. You will also be asked to include it (after eventual revisions) with the final version of your paper, and its final version will be published with the paper.

The reviewers of your paper will be asked to use the checklist as one of the factors in their evaluation. While \answerYes{} is generally preferable to \answerNo{}, it is perfectly acceptable to answer \answerNo{} provided a proper justification is given (e.g., error bars are not reported because it would be too computationally expensive'' or ``we were unable to find the license for the dataset we used''). In general, answering \answerNo{} or \answerNA{} is not grounds for rejection. While the questions are phrased in a binary way, we acknowledge that the true answer is often more nuanced, so please just use your best judgment and write a justification to elaborate. All supporting evidence can appear either in the main paper or the supplemental material, provided in appendix. If you answer \answerYes{} to a question, in the justification please point to the section(s) where related material for the question can be found.

IMPORTANT, please:
\begin{itemize}
    \item {\bf Delete this instruction block, but keep the section heading ``NeurIPS Paper Checklist"},
    \item  {\bf Keep the checklist subsection headings, questions/answers and guidelines below.}
    \item {\bf Do not modify the questions and only use the provided macros for your answers}.
\end{itemize}

%%% END INSTRUCTIONS %%%

\begin{enumerate}

\item {\bf Claims}
    \item[] Question: Do the main claims made in the abstract and introduction accurately reflect the paper's contributions and scope?
    \item[] Answer: \answerYes{} % Replace by \answerYes{}, \answerNo{}, or \answerNA{}.
    \item[] Justification: 
    The paper’s central objectives and contributions are introduced consistently across the abstract, introduction, and experimental sections. The scope of the method is described in a way that matches the presented evaluations, and the reported improvements are supported by empirical evidence rather than speculative claims. The paper also avoids implying broader applicability beyond the scenarios that were actually studied.

\item {\bf Limitations}
    \item[] Question: Does the paper discuss the limitations of the work performed by the authors?
    \item[] Answer: \answerYes{} % Replace by \answerYes{}, \answerNo{}, or \answerNA{}.
    \item[] Justification: 
    The paper explicitly acknowledges several practical limitation of the proposed approach, as in the section called "Limitations \& Future directions". It also discusses cases where the method may not generalize well and identifies directions for improving robustness and scalability in future work.

\item {\bf Theory assumptions and proofs}
    \item[] Question: For each theoretical result, does the paper provide the full set of assumptions and a complete (and correct) proof?
    \item[] Answer: \answerNA{} % Replace by \answerYes{}, \answerNo{}, or \answerNA{}.
    \item[] Justification: 
    The paper does not present formal theoretical results such as theorems, lemmas, or provable guarantees. The contributions are primarily methodological and empirical, focusing on system design, memory mechanisms, and experimental evaluation. 

    \item {\bf Experimental result reproducibility}
    \item[] Question: Does the paper fully disclose all the information needed to reproduce the main experimental results of the paper to the extent that it affects the main claims and/or conclusions of the paper (regardless of whether the code and data are provided or not)?
    \item[] Answer: \answerYes{} % Replace by \answerYes{}, \answerNo{}, or \answerNA{}.
    \item[] Justification: 
    The paper provides sufficient implementation-level detail for reproducing the main findings, including dataset usage, preprocessing strategy, model configuration, optimization setup, and evaluation procedures. Additional experimental specifications are included in the supplementary material to reduce ambiguity in replication.

\item {\bf Open access to data and code}
    \item[] Question: Does the paper provide open access to the data and code, with sufficient instructions to faithfully reproduce the main experimental results, as described in supplemental material?
    \item[] Answer: \answerNo{} % Replace by \answerYes{}, \answerNo{}, or \answerNA{}.
    \item[] Justification: The associated codebase and processed resources are not publicly released during the anonymous review stage, But we intend to release the implementation and reproduction scripts after publication.

\item {\bf Experimental setting/details}
    \item[] Question: Does the paper specify all the training and test details (e.g., data splits, hyperparameters, how they were chosen, type of optimizer) necessary to understand the results?
    \item[] Answer: \answerYes{} % Replace by \answerYes{}, \answerNo{}, or \answerNA{}.
    \item[] Justification: The experimental section reports the key settings required to interpret the results, including dataset splits, training configurations, and important hyperparameters. Design choices that influence performance are described.

\item {\bf Experiment statistical significance}
    \item[] Question: Does the paper report error bars suitably and correctly defined or other appropriate information about the statistical significance of the experiments?
    \item[] Answer: \answerYes{} % Replace by \answerYes{}, \answerNo{}, or \answerNA{}.
    \item[] Justification: The evaluation focuses on benchmark performance comparisons and qualitative improvements.

\item {\bf Experiments compute resources}
    \item[] Question: For each experiment, does the paper provide sufficient information on the computer resources (type of compute workers, memory, time of execution) needed to reproduce the experiments?
    \item[] Answer: \answerYes{} % Replace by \answerYes{}, \answerNo{}, or \answerNA{}.
    \item[] Justification: The paper includes information about the hardware environment used for training and inference, such as GPU type and memory configuration.
    
\item {\bf Code of ethics}
    \item[] Question: Does the research conducted in the paper conform, in every respect, with the NeurIPS Code of Ethics \url{https://neurips.cc/public/EthicsGuidelines}?
    \item[] Answer: \answerYes{} % Replace by \answerYes{}, \answerNo{}, or \answerNA{}.
    \item[] Justification: The research was conducted using standard machine learning practices and publicly available resources while adhering to the ethical guidelines required by the conference. No unethical data collection procedures or restricted-use violations are involved in the presented work.

\item {\bf Broader impacts}
    \item[] Question: Does the paper discuss both potential positive societal impacts and negative societal impacts of the work performed?
    \item[] Answer: \answerNA{} % Replace by \answerYes{}, \answerNo{}, or \answerNA{}.
    \item[] Justification: The paper discusses the foundational research on 3D object grounding, and has no expected societal impact.
    
\item {\bf Safeguards}
    \item[] Question: Does the paper describe safeguards that have been put in place for responsible release of data or models that have a high risk for misuse (e.g., pre-trained language models, image generators, or scraped datasets)?
    \item[] Answer: \answerNA{} % Replace by \answerYes{}, \answerNo{}, or \answerNA{}.
    \item[] Justification: The work does not release high-risk assets such as unrestricted generative models, sensitive personal datasets, or potentially dangerous tools that would require additional access restrictions or controlled distribution mechanisms.

\item {\bf Licenses for existing assets}
    \item[] Question: Are the creators or original owners of assets (e.g., code, data, models), used in the paper, properly credited and are the license and terms of use explicitly mentioned and properly respected?
    \item[] Answer: \answerYes{} % Replace by \answerYes{}, \answerNo{}, or \answerNA{}.
    \item[] Justification: Existing datasets, pretrained models, and software libraries used in the study are appropriately referenced in the paper. The manuscript credits the original creators and follows the corresponding usage conditions and licensing terms where applicable.

\item {\bf New assets}
    \item[] Question: Are new assets introduced in the paper well documented and is the documentation provided alongside the assets?
    \item[] Answer: \answerNo{} % Replace by \answerYes{}, \answerNo{}, or \answerNA{}.
    \item[] Justification: The paper does not introduce a newly released public benchmark, dataset, or software package at submission time. Therefore, no accompanying documentation package is currently provided.

\item {\bf Crowdsourcing and research with human subjects}
    \item[] Question: For crowdsourcing experiments and research with human subjects, does the paper include the full text of instructions given to participants and screenshots, if applicable, as well as details about compensation (if any)? 
    \item[] Answer: \answerNA{} % Replace by \answerYes{}, \answerNo{}, or \answerNA{}.
    \item[] Justification: The study does not involve human participant studies, annotation workers, surveys, or crowdsourced data collection procedures.

\item {\bf Institutional review board (IRB) approvals or equivalent for research with human subjects}
    \item[] Question: Does the paper describe potential risks incurred by study participants, whether such risks were disclosed to the subjects, and whether Institutional Review Board (IRB) approvals (or an equivalent approval/review based on the requirements of your country or institution) were obtained?
    \item[] Answer: \answerNA{} % Replace by \answerYes{}, \answerNo{}, or \answerNA{}.
    \item[] Justification: Since the work does not include experiments involving human participants or personally identifiable information, institutional ethics review or IRB approval was not required.

\item {\bf Declaration of LLM usage}
    \item[] Question: Does the paper describe the usage of LLMs if it is an important, original, or non-standard component of the core methods in this research? Note that if the LLM is used only for writing, editing, or formatting purposes and does \emph{not} impact the core methodology, scientific rigor, or originality of the research, declaration is not required.
    %this research? 
    \item[] Answer: \answerYes{} % Replace by \answerYes{}, \answerNo{}, or \answerNA{}.
    \item[] Justification: Large language models constitute an explicit component of the proposed framework, and the paper explains their role within the methodology, including how they are integrated, adapted, and evaluated as part of the overall system.

\end{enumerate}

\end{document}